%% file: main.tex
\documentclass[11pt,a4paper]{article}
\usepackage[hyperref]{emnlp2020}
\usepackage{times}
\usepackage{latexsym}
\usepackage{url}

\usepackage{array}
\usepackage{xcolor}
\usepackage{makecell}
\usepackage{graphicx}
\usepackage{multirow}
\usepackage{microtype}

\newcolumntype{P}[1]{>{\centering\arraybackslash}p{#1}}

\usepackage{microtype}

\aclfinalcopy 

\title{What if we had no Wikipedia? Domain-independent Term Extraction from a Large News Corpus}

\author{Yonatan Bilu \\
  IBM Research AI \\
  \small{\texttt{yonatanb@il.ibm.com}} \\\And
  Shai Gretz \\
  IBM Research AI \\
  \small{\texttt{avishaig@il.ibm.com}} \\\And
  Edo Cohen \\
  IBM Research AI \\
  \small{\texttt{edo.cohen@ibm.com}} \\\And
  Noam Slonim \\
  IBM Research AI \\
  \small{\texttt{noams@il.ibm.com}} 
  }

\date{}

\begin{document}
\maketitle
\input{abstract.tex}

\input{intro.tex}

\input{related_work.tex}

\input{data.tex}

\input{observations}

\input{candidates_classification}

\input{ambiguity_detection}

\input{clustering}

\input{candidate_classification_further}

\input{discussion}


\bibliography{main}
\bibliographystyle{acl_natbib}

\end{document}


\maketitle
\section{Context Homogeneity}
One way to try and deduce the role of extracted candidates is to examine the context in which they appear. Specifically, we suggest looking at a set of such contexts in unison, and measure its homogeneity or heterogeneity. We postulate that AWTs will tend to appear in more heterogeneous contexts than WT which are NAWTs. 

Moreover, wondering who are the 
noun phrases which are common in LNCS but absent from Wikipedia, we observe that tend to domain-specific, relating to some narrow field covered in LNC. Quite possibly, it is this specificity which is the reason why they were not included in Wikipedia despite their apparent abundance (examination of such terms suggest that many of them are specific economic terms). Hence context homogeneity may also be relevant to discerning between WTs and not WTs in our dataset.

We define \textit{context variance}, a measure the heterogeneity of a set of sentences, as follows. Let $c$ be a candidate term. Let $S = \{s_1, \ldots, s_k\}$ be a sample of sentences in which it appears. Let $U = \{u_1, \ldots, u_m\}$ be the collection of unigrams appearing in these sentences (excluding $c$). 

Denote 
$$n_i = |\{s \in S : u_i \in S\}|$$ and $$p_i= \frac{n_i}{\Sigma_j n_j}$$
Then the \textit{context variance} of a term $c$, $CV(c)$, is deined as the variance of the distribution defined by the $p_i$s. 

Accordingly, we expect that if $c$ is an AWT, $CV(c)$ will be relatively higher than when it is a ``regular" WT. Similarly, that when a $c$ is an WT, $CV(c)$ will be relatively higher than when it is a non-WT.

\section{Error Analysis}
\subsection{Candidate classification vs. Wikipedia}
Examining the errors made by the fine-tuned BERT model over extracted bigrams, suggests that false positives tend to be associated with the financial domain. The top scoring non-WTs are: \textit{financial information}, \textit{financial performance}, \textit{total debt}, \textit{commodity prices} and \textit{banking sector}. This perhaps highlights the fact that to some extent, the decision of which terms to include in Wikipedia and what terminology to use may be arbitrary, as the term \textit{commodity prices index} does appear in Wikipedia, but at the time of writing the term \textit{commodity prices} does not, not even as a co-redirect.

Conversely, the false-negatives seem to be associated with units and measures. The lowest scoring Wts are \textit{14 points}, \textit{9 months}, \textit{3 percent}, \textit{18 months} and \textit{average rating}. Hence, in future work, it might be worthwhile to define some filtering rules not only on extracted noun phrases, but also on the Wikipedia terms, which might exclude at least the first four of these false-negative terms.

\subsection{Candidate classification vs. ACL RD-TEC 2.0}
Examining the errors made by the fine-tuned BERT model on the ACL RD-TEC 2.0 benchmark \cite{qasemizadeh2016acl}, we see that the top scoring false-positives are \textit{sight-impaired users},
\textit{human users}, \textit{young men}, \textit{human interaction} and \textit{human subjects} (the latter three being WTs). Conversely, the lowest scoring false-negatives are \textit{boosting approach}, \textit{ranking function}, \textit{relative importance},
\textit{certainty conditions}, and \textit{relevant times}. The latter perhaps exemplifies that the ground truth labeling is somewhat subjective, and depends not only on the term itself, but also on the context in which it appears.

\subsection{Clustering Analysis}
To provide a better understanding of the behavior of the clustering methods, we examined relatively large clusters with low pair-wise precision, in the clustering of co-redirects of bigrams. A common error in the output of both GloVe-Har and GloVe-Agg is when clusters contain WTs that share a token, but actually redirect to different Wikipedia pages. For example, GloVe-Har clusters together \textit{Public work}, \textit{Public policy} and \textit{Public sector}, which redirect to three different Wikipedia pages. This can be explained by the use of the average GloVe vectors similarity measure, attaching these candidates by the common \textit{Public} term. 

For TF-sIB, a common error relates to similar context of different WTs. For example, the candidates \textit{Manchester Utd.}, \textit{Crystal Palace} and \textit{Jose Mourinho} are clustered together, as they all relate to English association football. To some extent, these errors occur when the context-based representation entails a categorical cluster, rather than a conceptual one.

\section{Methods}
\subsection{DeepSets}
In order to design an architecture which operates over sets we rely on the work of \citet{zaheer2017deep} that show that any function which takes sets as input can be universally approximated by the following decomposition $$f(X)=\rho\left(\sum_{i=1}^m\phi(x_i)\right)$$
where $X=\lbrace x_1,\dots,x_m\rbrace$ is a set, and $\rho, \phi$ are general transformations. Note that any function which takes the form above is inherently order invariant due to the \textit{sum} operator. In their paper, the authors experiment with other operators which are permutation invariant, such as the \textit{max} operator. 

In our setting, each n-gram induces a set vectors - the different contextual representation of the n-gram in the sentences in which it appears. Specifically, for each n-gram we sample $100$ sentences from LNC and extract the contextual representation of the n-gram from the last layer of BERT. We use BERT-base, where each token representation is a $768$ dimensional vector, concatenating $n$ such tokens results with $n\times 768$ dimensional vector which is a single contextual representation. In our experiment the configuration which worked best is setting $\phi$ as a 3-layered MLP, with input dimension of $768$, a single hidden layer with $20$ neurons and a single output. We then summed up these outputs, that is, $\rho$ just the identity function (or division by $n$ to scale the results). This is a somewhat  degenerate variant of DeepSet, but nonetheless surpasses some other variants we tried.

\bibliography{main}
\bibliographystyle{acl_natbib}

%% file: abstract.tex
\begin{abstract}
One of the most impressive human endeavors of the past two decades is the collection and categorization of human knowledge in the free and accessible format that is Wikipedia. In this work we ask what makes a term worthy of entering this edifice of knowledge, and having a page of its own in Wikipedia? To what extent is this a natural product of on-going human discourse and discussion rather than
an idiosyncratic choice of Wikipedia editors? Specifically, we aim to identify such ``wiki-worthy" terms in a massive news corpus, and see if this can be done with no, or minimal, dependency on actual Wikipedia entries. We suggest a five-step pipeline for doing so, providing baseline results for all five, and the relevant datasets for benchmarking them. Our work sheds new light on the domain-specific Automatic Term Extraction problem, with the problem at hand being a domain-independent variant of it.
\end{abstract}

%% file: intro.tex
\section{Introduction}
The impact of Wikipedia on modern life in general, and NLP research in particular, can not be overstated. It is hard to imagine our world without Wikipedia - yet in this work we ask 
the readers to suspend their 
disbelief and do just that. Suppose that 
instead of founding Bomis, Nupedia and ultimately Wikipedia, Jimmy Wales would have stayed on as a trader in Chicago Options Associates, and only now, in 2020, would have decided to establish a free online encyclopedia. How could modern NLP algorithms and resources have assisted him in doing so? 
As a first step, 
could they automatically, and with reasonable 
accuracy, suggest the terms and concepts that should compose Wikipedia? 
Clearly, one could have constructed a close approximation of Wikipedia by piecing together information from online encyclopedias, glossaries and dictionaries. So to refine our question - is it possible to reconstruct Wikipedia by analyzing current human discourse, and, specifically, as it is reflected in the media?

This question is closely related to Automatic Term Extraction (ATE), a line of research that can trace its roots back to \citet{luhn1957statistical}, who recognized the importance of identifying the terms in a document which will facilitate its coherent retrieval. Indeed, document-level term extraction has been one of the main tasks in Information Retrieval. However, closer to this work is domain-specific ATE, which, given a 
corpus of documents related to a well defined knowledge domain, aims to extract the salient terms fundamental to this domain. Work on ATE have been instrumental for facilitating downstream tasks such as indexing, mention detection \cite{usbeck2015gerbil},
extracting textual themes \cite{bawakid2015using}, lexicon construction \cite{velardi2008mining}, ontology learning \cite{brewster2007dynamic}, and knowledge organization \cite{chisholm2016discovering} -- all within the context of a well defined domain.  

Our setup is somewhat different. 
Given a large and diverse (newspaper) corpus, can we identify terms which represent 
titles of Wikipedia articles, 
and can we rank them by order of importance?

The automatic construction of Wikipedia entails a few sub-tasks, which we will elaborate on below. First and foremost, there is the task of \textbf{Candidate classification}. Given a term, e.g., \textit{Artificial intelligence}, we want to predict whether it is worthy of being part of Wikipedia or not. But appearance in Wikipedia is not the only distinction we should make. For example, consider the term \textit{State}. When one searches for \textit{State} in Wikipedia, she may be interested in the political structure (such as the state of New York) or to the state of a computer system. These Wikipedia titles differ from regular pages; thus, we define a second task, \textbf{Ambiguity detection}, whose goal is to identify when a Wikipedia term should be a Wikipedia disambiguation page or a regular page. In addition, most Wikipedia terms do not have a single way of mentioning them in writing. For referring to \textit{Artificial intelligence}, for example, one may use the term \textit{Artificial Intelligence}, \textit{AI}, or even \textit{Cognitive systems}. Thus, we define a third task, \textbf{Surface form clustering}, whose goal is to cluster together terms which should direct to the same Wikipedia article.

For our suggested pipeline, we also include a preliminary step, \textbf{Candidate extraction}, whose goal is to extract terms from a large corpus which are likely to appear in Wikipedia, and a concluding step, \textbf{Term ranking}, for ranking terms (or clusters of terms) according to importance, to facilitate an appropriately priortized authoring of Wikipedia pages. The full pipeline is depicted in Figure \ref{fig:arch}.


\input{sys_arch_fig.tex}


We describe baseline results for all steps, but focus on \textit{candidate classification}, \textit{ambiguity detection} and \textit{surface forms clustering}, of which the latter two, due to the domain-specific nature of previous 
work on ATE, so far 
received little attention (as far as we know, \textit{ambiguity detection} in this context is a novel task). 
In addition, we examine how well humans can perform the \textit{candidate classification} step, and how the models we developed in a domain-independent manner, can lead to an effective baseline in a domain-specific context.

In subsequent sections we use the following terminology. We say that an n-gram 
is a \textit{Wikipedia Term} (WT) if it corresponds to a title of a Wikipedia page, or redirects to one, 
and that a WT is an \textit{Ambiguous Wikipedia Term} (AWT) if the 
Wikipedia page it is associated with is a disambiguation page. Conversely, we say that a WT is 
\textit{Non-Ambiguous Wikipedia Term} (NAWT) if it is not an AWT.
We say that two WTs 
are co-redirects if they are associated with the same Wikipedia page,
i.e., correspond to the same Wikipedia article. 
In this terminology, Step 2 discerns between WTs and non-WTs; 
Step 3 discerns between AWTs and NAWTs; 
and the clustering of Step 4 aims to cluster together two WTs iff they are co-redirects. 
Finally, we will also be interested in the relations between WTs and domain-specific terms, denoted here as DSTs.

%% file: sys_arch_fig.tex
\begin{figure}[t]
\centering
\includegraphics[width=75mm]{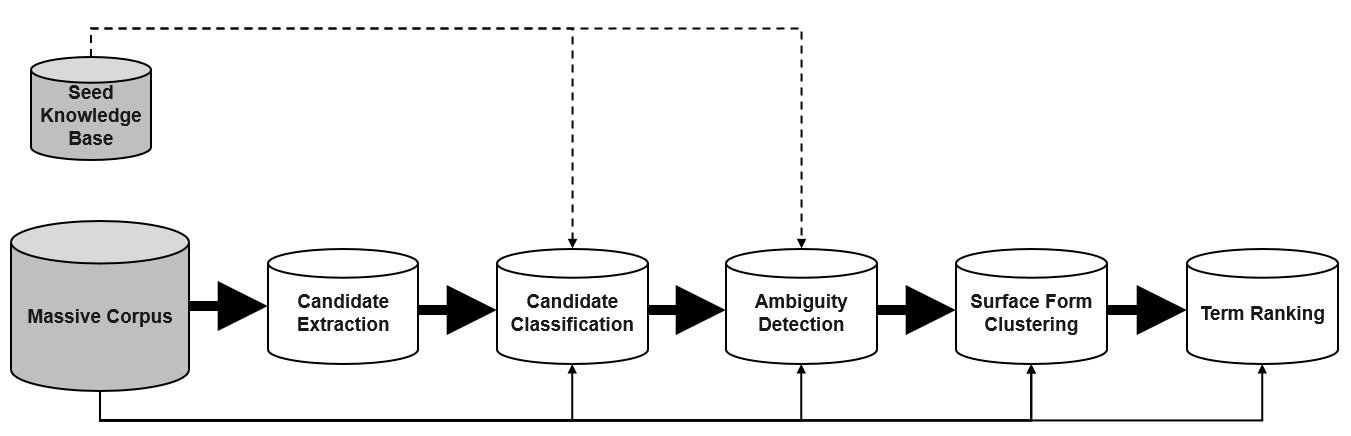}
\caption{
An overview of the suggested pipeline. A seed knowledge base is needed for the supervised classification methods described for \textit{candidate classification} and \textit{ambiguity detection}.\label{fig:arch}}
\end{figure}

%% file: related_work.tex
\section{Related Work}
Interest in Automatic Term Extraction has initially been motivated by Information Retrieval needs - to identify salient terms at the document level \cite{luhn1957statistical}. By contrast, in the field of \textit{terminology}, the focus of ATE was oriented toward the corpus as a whole. Such research started at least as early as the early 90's \cite{auger1991automatisation}, which led 
to real-world solutions 
for this task like TERMINO \cite{plante1998depoulliment} and LEXTER \cite{bourigault1992surface}. See \cite{castellvi2001automatic,zhang2008comparative} 
for a review of these earlier works and systems. As with other fields 
in NLP, initially such systems were mostly rule-based, with later works introducing machine-learning techniques (e.g., \citealp{conrado2013machine}).

By and large, work on ATE focuses on the terminology of specific, well-defined domains, such as the medical domain \cite{fraser2019extracting, kim2003genia, lossio2016biomedical}, natural language processing \cite{qasemizadeh2016acl}, and information technology services \cite{mohapatra2018domain}. In the context of scientific literature ATE is sometimes used as a means toward summarizing scientific papers. For such uses, one is not interested merely in terms specific to the domain, but in those which describe specific aspects of the paper, such as terms which describe the \textit{techniques}, \textit{focus} and \textit{domains} \cite{gupta2011analyzing} of the paper, and also the \textit{applications} discussed therein \cite{tsai2013concept}. This latter work also identifies the needs to cluster together different surface forms which refer to the same technique or application. They suggest a clustering algorithm based on juxtaposed citation indices, a technique specific to scientific literature. The need for clustering of surface forms is also discussed in \citet{peng2016event}, where the terms of interest are mentions of events.

Domain-specific solutions, which are probably indeed of greater practical interest than the domain-independent approach, tend to face the following challenges - (a) the corpus is usually of moderate size; 
(b) evaluation tends to be challenging and requires manual annotation; (c) solutions tend to be domain-specific, with no single method arising as ``best practice" in the field \cite{zhang2016jate, zhang2018adapted}; (d) domain-specific terms need to be discerned from domain-independent terms. Hence, we hope that it might be of interest to complement domain-specific ATE with a domain-independent one.


%% file: data.tex
\section{Data Preparation}\label{sec:data}
The data examined for candidate extraction comes from a massive English news articles corpus provided by LexisNexis.\footnote{\url{https://www.lexisnexis.com/en-us/home.page}} This corpus contains 
around $400$ millions articles, from which we 
(uniformly) 
sampled at random $1$ million distinct sentences. 
During sampling we discarded sentences shorter than $10$ tokens and sentences containing a newline, and kept only one copy of duplicate texts. We hereafter refer to the entire corpus as \textit{LNC} (LexisNexis Corpus), and to the sample as \textit{LNCS}. 

Each sentence was analyzed with the SpaCy \cite{honnibal2015improved} parser, and the noun phrases therein were extracted. Leading stop-words were removed from these noun phrases, and 
noun phrases containing only stop-words were discarded.
We then kept only unigrams, bigrams, and trigrams, converting them to lowercase form.

Each n-gram was labeled for being a WT -- or not -- by querying English Wikipedia (online)
with both the lowercase version of its surface form, and with all tokens capitalized. 
An n-gram identified as a WT was further marked 
as an AWT if the corresponding Wikipedia article contains the hallmark text ``This disambiguation page lists articles associated with" (and as an NAWT, otherwise).

For unigrams and bigrams a threshold of appearing in at least $50$ sentences was set, and for trigrams a threshold of $10$. Table \ref{tab:data} lists the number of candidates so extracted, and 
the number of WTs and AWTs among them. Note that co-redirects are considered in this table, and during the initial $3$ steps of the pipeline, as distinct WTs, with the objective of the \textit{surface forms clustering} step being to group them together. 

We denote the set of collected noun phrases as LNNP (LexisNexis Noun Phrases), and define subsets of it by $LNNP_n = \{x \in LNNP : \textrm{x is an n-gram}\}$, and $LNNP_S = \bigcup_{n \in s} LNNP_n$. 

\input{data_table.tex}



Table \ref{tab:non-WT} lists bigrams and trigrams which are frequent in LNCS, but are {\it not\/} 
WTs. As can be seen, 
all are phrases which do appear in Wikipedia. The examples depict the common reasons for this - being domain-specific (e.g., \textit{forward-looking statements}) or being a general phrase
(e.g. \textit{tens of thousands}).

\input{examples_table.tex}

For the purpose of exploring \textit{surface forms clustering} the collected data is too sparse. That is, among the collected n-grams there are very few examples of co-redirects - mostly just singular and plural forms of the same term. Moreover, this step in only relevant for WTs, and more specifically for NAWTs. Hence, for the analysis associated with this step only, we first define:
$LNNP^*_n$ as the subset of $LNNP_n$ where elements are NAWTs appearing in at least  100 
LNC sentences. We then augment each such set with all surface forms that are co-redirects of one of the n-grams therein, provided that they, too, appear in at least 100 
LNC sentences. It should be noted that when augmenting co-redirects, there is no limit on their number of tokens. For example, co-redirects augmenting unigram candidates may (and do) contain 
multiple tokens. We denote such an augmentation of $LNNP^*_n$ as $LNNP^{aug}_n$.

Finally, for examining the relevance of domain-independent ATE to domain-specific ATE, 
we further considered the ACL RD-TEC 2.0 benchmark \cite{qasemizadeh2016acl}. This benchmark lists $300$ abstracts of papers from the ACL anthology, manually annotated for terms consisting 
of specialized vocabulary related to NLP. We extracted all noun phrases from these abstracts, and all annotated terms. We define a noun phrase as a DST for this benchmark if it is identical (ignoring case and leading stop-words) to one of the manually annotated terms.

All the above datasets will be made freely available upon publication of this work.

%% file: data_table.tex
\begin{table}[ht]
\small
  \begin{center}
    \begin{tabular}{|l|c|c|c|} 
  \hline
      \textbf{Subset} & \textbf{\# Cand.} & \textbf{\# WT} & \textbf{\# AWT}\\
      \hline
      $LNNP_1$ & 6860 & 6306 & 2603 \\
      \hline
      $LNNP_2$ & 1630 & 1075 & 120\\
      \hline
      $LNNP_3$ & 1315 & 663 & 20 \\
      \hline
    \end{tabular}
    \caption{Number of ngrams extracted from LNCS, and the number of WTs and AWTs. Note that all AWTs are also WTs, and are included in that count.}
    \label{tab:data}
  \end{center}
\end{table}
\vspace{-0.5cm}

%% file: examples_table.tex
\begin{table}[ht]
\small
  \begin{center}
    \begin{tabular}{|l|c|c|c|} 
  \hline
      \textbf{Phrase} & \textbf{LNCS} & \textbf{LNC} & \textbf{Wiki}\\
      \hline
      fourth quarter & 1234 & 9.3M & 12164\\
      \hline
      publication name & 737 & 4.6M & 139 \\
      \hline
      forward-looking  & 2476 & 17.8M & 34 \\
      statements & & & \\
      \hline
      tens of thousands & 244 & 1.7M & 13208 \\
      \hline
      relative price change & 152 & 1.1M & 3 \\
      \hline
    \end{tabular}
    \caption{Examples of noun phrases which are common in LNCS, but are not WTs. Columns detail the number of sentences in which they appear in each of the three corpora. Note that for LNC numbers listed are in units of millions.}
    \label{tab:non-WT}
  \end{center}
\end{table}
\vspace{-0.5cm}

%% file: observations.tex
\section{Preliminary Observations}\label{sec:observations}
In data preparation we followed the common wisdom of ATE research, and extracted noun phrases which are relatively common in the corpus (cf. \citealp{castellvi2001automatic}). As can be seen in Table \ref{tab:data}, this already achieves the stated goal of the \textit{candidate extraction} step, as $LNNP$ is abundant with WTs. Hence, for the purpose of this work, we conclude that this simple method already achieves a reasonable baseline, and do not explore this step further.

Furthermore, Table \ref{tab:data} shows that $LNNP_1$ is composed of nearly only WTs, with a roughly even split between AWT and NAWTs. Conversely, $LNNP_{2,3}$ contain nearly no AWTs, with a roughly equally split between WTs and non-WTs. Hence, to address \textit{candidate classification} we restrict our analysis to $LNNP_{2,3}$, while for \textit{ambiguity detection} we restrict our analysis to $LNNP_1$.

Finally, we observe that one can also attain reasonably good results for \textit{term ranking} based on frequency. Specifically, a commonly used metric for defining the ``importance" of a Wikipedia article is the number of other articles which link to it (though many other metrics exist, e.g., \citet{thalhammer2016pagerank, lewoniewski2016quality}). We find that this metric is strongly correlated with the frequency of the article's title in LNC (Spearman rank correlation $0.77$). Hence, for the purpose of obtaining baseline results for the suggested pipeline, we find this simple technique adequate, and, as with \textit{candidate extraction}, do not explore this step further.

%% file: candidates_classification.tex
\section{Candidates Classification}\label{sec:cand_class}
As described above, the \textit{candidates classification} task is to determine which of the n-grams in $LNNP_{2,3}$ are WTs. We examine both supervised and unsupervised methods for this:

\paragraph{Relative frequency (RF):} Following \citet{nakagawa2002simple}, we computed 
for each n-gram the (log) ratio between its frequency in the corpus, and the product of the frequency of its constituent unigrams. This can be seen as a refinement of the initial \textit{candidate selection} stage, and as an estimate for its \textit{termhood} (cf. \citealp{kageura1996methods}).

\paragraph{Context variance (CV):} For a given n-gram, we aimed to measure the heterogeneity of the contexts in which it appears by sampling 1000 sentences containing it from LNC, and then computing the 
variance of the unigram-distribution over them - full details are in the appendix.

\paragraph{BERT:} We fine-tuned BERT directly on the surface forms of the n-grams in the usual way: Data was split into train, development and evaluations sets\footnote{The size of the sets was set to 20\%, 20\% and 60\% of the data, respectively. This reflects a real-world scenario where we are given a small set of labeled terms, and aim to predict over a large set.}, and the BERT model was fine tuned on the former and evaluated on the latter. Importantly, Our data sometimes contains both singular and plural forms of a noun phrase. To prevent the model from simply ``copying" the label of one form in the training set to its matching form in the evaluation set, for each such singular-plural pair we discarded the less LNCS-frequent one from the analysis.

Table \ref{tab:unsupervised} reports the Spearman rank correlation of RF and CV measures with the ground-truth labels. In addition, we compute accuracy by considering the top $k$ scoring n-grams as positive examples, 
and the remainder as negative, with $k$ being the true number of positives. As ground truth, we consider NAWTs as positive examples, and non-WTs as negative examples. Accordingly, for CV, we consider as ``top scoring" those candidates which induce lower variance, as we associate low variance with being a WT (hence the negative values in Table \ref{tab:unsupervised};  see appendix).

\input{unsupervised_table.tex}

Results suggest that these unsupervised methods are correlated with the ground truth 
with the expected sign (see appendix), 
but moderately so. This leads to an accuracy which is higher than the trivial baseline of predicting the majority class, especially in the case of $LNNP_3$. Interestingly, we also measured the accuracy of simply predicting the $k$ most common candidates to be WTs. This yields an accuracy that is actually lower than the majority baseline ($0.55$) in the case of $LNNP_2$, and essentially the same as this baseline ($0.50$) in the case of $LNNP_3$.

As shown in Table \ref{tab:supervised} fine tuning BERT on a small set of bigrams and trigrams (independently) yielded a much higher accuracy than the unsupervised methods. A supervised approach is not completely aligned with the question we asked at the onset of this work - how to identify WTs \textit{de-novo}, but might be interesting in a scenario where experts identify a small number of ``seed" terms as WTs, and this set is then expanded by -- or with assistance of -- automatic means.
 
\input{bert_table.tex}

With the proven success of BERT on so many NLP tasks,
one might not be surprised by the attained high accuracy. Yet, keep in mind that part of BERT's training data comes from Wikipedia. In particular, one might suspect that, as a language model, BERT assigns higher probability to bigrams and trigrams 
WTs, since they might be over represented in its training data relative to non WTs.

To try and control for this, we examined whether the number of Wikipedia sentences in which an n-gram appears in is a good predictor for it being a WT. We ranked the evaluation set according to this frequency, and predicted that the top $k$ n-garms are WT and the remainder are not (with $k$ being the number of WTs in the set). For bigrams, this yields an accuracy of $0.69$, and for trigrams an accuracy of $0.79$. In both cases this is higher than the majority baseline, but falls short of the BERT-based accuracy. This suggests that frequency alone can not account for this success, though further attention should be given in future work to the apparent different gap magnitude between bigrams and trigrams.  


Analysis of the errors made by the fine-tuned BERT model seem to be associated with specific semantic categories. Details 
are in the appendix. 

%% file: unsupervised_table.tex
\begin{table}[ht]
\small
  \begin{center}
    \begin{tabular}{|l|c|c|c|} 
  \hline
      \textbf{Dataset} & \textbf{Base} &
      \textbf{RF} & \textbf{CV} 
     \\
      \hline
      $LNNP_2$ & 0.59 &  0.65 (0.3) &  0.64 (-0.26)\\
      \hline
      $LNNP_3$ & 0.49 & N/A & 0.65 (-0.4) \\
      \hline
    \end{tabular}
    \caption{Accuracy and correlation (in parenthesis) to ground truth of unsupervised methods. \textit{Base} is the accuracy of the majority class baseline.}
    \label{tab:unsupervised}
  \end{center}
\end{table}

%% file: bert_table.tex
\begin{table}[ht]
\small
  \begin{center}
    \begin{tabular}{|l|c|c|c|c|} 
  \hline
      \textbf{Dataset} & \textbf{\# Train} & \textbf{\# Eval} & \textbf{Base} &
      \textbf{BERT} 
     \\
      \hline
      $LNNP_2$ & 299 & 870 & 0.63 & 0.82 \\
      \hline
      $LNNP_3$ & 258 & 792 & 0.49 & 0.82 \\
      \hline
    \end{tabular}
    \caption{Size of training and evaluation sets for fine-tuned BERT classification; accuracy of majority baseline and BERT-based classification on the latter.}
    \label{tab:supervised}
  \end{center}
\end{table}
\vspace{-0.5cm}

%% file: ambiguity_detection.tex
\section{Ambiguity Detection}\label{sec:amb_det}
As described above, the \textit{ambiguity detection} task is to determine which of the n-grams in $LNNP_{1}$ are AWTs, and which are NAWTs (we discard non-WTs). We applied the same unsupervised methods described in Section \ref{sec:cand_class}, but to no avail. The CV method displayed low correlation with the ground truth (Spearman rank correlation $0.16$), leading to an accuracy of $0.61$ compared to a baseline of $0.59$.

Using BERT\footnote{Splitting $LNNP_1$ to train-dev-test with 0.2, 0.2 and 0.6 of the data, respectively.} lead to mixed results. In 16 of 20 runs, it failed to learn a meaningful model, and simply predicted the majority class. This led us to seek a more robust model, by leveraging the contexts in which a term appears.
To do do this we employed and architecture similar to Deep Set \cite{zaheer2017deep}:
For each WT we extract from LNC 100 sentences in which it appears. We then process the sentences with BERT, and extract the term's contextual vector representation . 
Each of these representations is a single training example used to train a fully-connected neural net. 
During evaluation, each of the $100$ sentences extracted for a WT 
is classified by the model, and the predicted label for a term is determined by the average scores for these sentences (see appendix for full details).

As can be seen in Table \ref{tab:unigrams}, this DeepSet method, which aggregates together the different contexts in which a WT appears, did manage to surpass the majority baseline in all runs, and seems somewhat better than the more naive application of BERT. This suggests that the contexts in which a term appears, and perhaps also the relations between them, are indeed related to whether or not it is an AWT.

\input{unigrams_table.tex}

%% file: unigrams_table.tex
\begin{table}[ht]
\small
  \begin{center}
    \begin{tabular}{|l|c|c|c|} 
  \hline
      \textbf{Method} & \textbf{\# fail} & \textbf{mean acc.} & \textbf{max acc.} 
     \\
      \hline
      BERT & 16 & $0.713 \pm 0.015$ & 0.729 \\
      \hline
      DeepSet & 0 & $0.725 \pm 0.005$ & 0.735 \\
      \hline
    \end{tabular}
    \caption{Classification results for discerning AWTs from NAWTs (from among WTs), over 20 runs. The model is considered to have failed to learn if its accuracy is no better than the majority-class baseline (0.61). Mean accuracy and standard deviations are computed over the runs which did not fail.}
    \label{tab:unigrams}
  \end{center}
\end{table}
\vspace{-0.5cm}

%% file: clustering.tex
\section{Surface Forms Clustering}\label{sec:surf_form_clust}
As described above, the \textit{ambiguity detection} task is to cluster $LNNP^{aug}_n$ so that two surface forms are in the same cluster iff they are co-redirects. We consider a $2$-phase approach, 
where the first phase is a rule-based clustering (RBC) which identifies candidates that should initially be merged. These rules have high precision, but identify only a relatively small number of candidates that should be merged. In the second phase a more standard clustering algorithm is used, alongside a more general similarity measure. This can be calibrated to yield a desired number of clusters.

Specifically, the RBC phase was done by merging candidates if they share common word-forms,\footnote{Using the python package: \url{https://github.com/gutfeeling/word_forms}} ignoring case, punctuation, space, and a trailing \verb|s| character. For example, \textit{Artificial-intelligence}, \textit{Artificial Intelligence} and \textit{Artificially intelligent}, co-redirects of the Wikipedia page \textit{Artificial intelligence}, are clustered together by this phase.


In the second phase, we examined four 
possible approaches for clustering candidates: 

\paragraph{GloVe-Agg and GloVe-Har:} Each candidate is represented by the average of the GloVe embeddings of its constituent unigrams \cite{pennington2014glove}. After merging candidates together in the RBC step, the vector representation of this initial cluster is the average of the vectors of all merged candidates. These vectors are then clustered based on their cosine similarity. In one variant we use Agglomerative clustering\footnote{Using the scikit-learn package: \url{https://scikit-learn.org}} to do this, and in the other Hartigan's K-Means \cite{hartigan1975,Slonim18297}. 

\paragraph{TF-sIB:} For each candidate, $100$ LNC sentences in which it appears are retrieved (if several candidates were merged during RBC, one is chosen arbitrarily). The information gain of each token in these sentences w.r.t the candidate surface form is computed, and the top $2000$ tokens are taken as features for a term frequency vector representation. The motivation for this approach is that co-redirects with different surface forms are likely to appear in similar contexts. These vectors are clustered using the Sequential Information Bottleneck (sIB) algorithm \cite{Slonim2002}.

\paragraph{BERT-Har:} For each candidate, $5$ LNC sentences in which it appears are retrieved (as above). For each candidate and respective sentence, we calculate the average BERT contextual token embeddings of the candidate's constituent unigrams, from the second to last layer of BERT.\footnote{In case a unigram is split to multiple word pieces, we calculate the average of its word pieces, up to 6 word pieces.} We then average these vectors across the 5 sentences to obtain a single candidate representation. The rest of this approach is similar to the approach using GloVe embeddings. We use Hartigan's K-Means for clustering, as it seemed to work somewhat better than Agglomerative clustering.

In all approaches we set the number of clusters to be the number of ground-truth Wikipedia titles. In addition, we filter about $1\%$ of the candidates for which one of the constituent unigrams is not found in GloVe's vocabulary, or was split to word pieces by BERT's tokenizer in a way that we were not able to merge back to a single candidate (e.g., if a candidate contained many `.' symbols).
The number of co-redirects and desired clusters, before and after the RBC step, is summarized in Table \ref{tab:clusteringData}.

\input{clustering_data_table.tex}

Table \ref{tab:clusteringResults} assesses 
the quality of the resulting clusters using two measures: adjusted rand index (ARI) and BCubed-F1 \cite{Amigo2009}. The best results 
are obtained by the BERT-Har approach, by a considerable margin.

\input{clustering_results.tex}

We compared a sample of generated clusters of $LNNP^{aug}_2$ for the BERT-Har and GloVe-Har methods. From this examination, the effect of the contextual representation is clear, especially for ambiguous tokens. For example, the token \verb|common| is part of economic WTs such as \verb|Common stock|, as well as more abstract WTs such as \verb|Common sense|. In the output of GloVe-Har, these two WTs are found in the same cluster, as can be seen at the lower cluster of Table \ref{tab:clusteringComparison}. The similarity between the two WTs is presumably a result of the shared token \verb|common| having the same GloVe representation. However, with BERT-Har, these terms are in distinct clusters. Moreover, as can be seen at the upper cluster of the table, the candidate \verb|Common stock| resides correctly with \verb|Equity shares|, as they are both co-redirects to the Wikipedia title \verb|Common stock|, even though they do not share a common token.

Where BERT-Har tends to fail is when the ground-truth resolution of Wikipedia co-redirects is too fine-grained. For example, \verb|Common stock| and \verb|Equity shares| are clustered together with the WT \verb|Company stock|, which is a redirect to a different Wikipedia title, \verb|Stock|. These subtleties are difficult to capture with the current methodology, and we leave handling them for future work. Moreover, it is not even clear whether for downstream applications, such a fine-grained distinction is desired or beneficial.



\input{clustering_examples_table.tex}

%% file: clustering_data_table.tex
\begin{table}[ht]
\small
  \begin{center}
    \begin{tabular}{|l|P{1.6cm}|P{1.6cm}|P{1.4cm}|} 
  \hline
      \textbf{Dataset} & \textbf{\# Before RBC} & \textbf{\# After RBC} & \textbf{\# Clusters}\\
      \hline
      $LNNP^{aug}_1$ & $3376$ & $2382$ & $500$ \\
      \hline
      $LNNP^{aug}_2$ & $9387$ & $5350$ & $987$\\
      \hline
    \end{tabular}
    \caption{Number of co-redirects for each n-gram before and after rule-based clustering, and number of clusters.}
    \label{tab:clusteringData}
  \end{center}
\end{table}

%% file: clustering_results.tex
\begin{table}[ht]
\small
  \begin{center}
    \begin{tabular}{|l|P{1.6cm}|P{1.0cm}|P{1.6cm}|} 
  \hline
      \textbf{Dataset} & \textbf{Method} & \textbf{ARI} & \textbf{BCubed-F1}\\
      \hline
       \multirow{3}{*}{$LNNP^{aug}_1$} & GloVe-Agg & $0.38$ & $0.5$\\
      \cline{2-4}
       & GloVe-Har & $0.37$ & $0.5$\\
      \cline{2-4}
       & TF-sIB & $0.4$ & $0.51$\\
      \cline{2-4}
       & BERT-Har & $\textbf{0.46}$ & $\textbf{0.58}$ \\
        \hline
      \multirow{3}{*}{$LNNP^{aug}_2$} & GloVe-Agg & $0.43$ & $0.56$ \\
        \cline{2-4}
       & GloVe-Har & $0.47$ & $0.57$ \\
      \cline{2-4}
       & TF-sIB & $0.46$ & $0.55$ \\
      \cline{2-4}
       & BERT-Har & $\textbf{0.53}$ & $\textbf{0.61}$ \\
      \hline
     
    \end{tabular}
    \caption{Adjusted rand index (ARI) and BCubed-F1 of four clustering methods. Best results for co-redirects of each n-gram are in bold.}
    \label{tab:clusteringResults}
  \end{center}
\end{table}

%% file: clustering_examples_table.tex
\begin{table*}[t]
\small
\begin{center}
\begin{tabular}{|P{2cm}|P{13cm}|}
\hline
\textbf{Method} & \textbf{Cluster of WTs} \\
\hline
\multirow{5}{*}{BERT-Har} &  Preferred equity, Preferred Equity, Convertible preferred stock, Preferred stocks, Preferred stock, Convertible Preferred Stock, Preferred Stock (Preferred stock) \\
&  Equity security, Capital stock, Equity securities, Company stock (Stock)  \\
& Equity shares, Common stocks, Common Stock, Common stock (Common stock)\\
& Share price, Stock price, Share prices (Share price)\\
& Common equity, Common Equity (Common equity)\\
\hline
\multirow{4}{*}{GloVe-Har} &  Common share, Common shares, Common stocks, Common stock, Common Stock (Common stock)\\
&  Paine's Common Sense, Common Sense, Thomas Paine's Common Sense (Common Sense (pamphlet))     \\
&  Common Ground, Common ground   (Common Ground)   \\
&  Common sense, Common-sense (Common sense)\\
\hline
\end{tabular}
\caption{Comparing between clusters containing the WT \textit{Common stock}, using BERT-Har (top) and GloVe-Har (bottom). An entire cell corresponds to a single generated cluster. In parenthesis: the ground-truth Wikipedia title for the respective WTs.}
\label{tab:clusteringComparison}
\end{center}
\end{table*}

%% file: candidate_classification_further.tex
\section{More on Candidate Classification}
\subsection{Human Performance}
The answer to the question of what makes a term ``Wikipedia worthy" is highly subjective, and depends on the views of Wikipedia editors. Hence, to appreciate the difficulty of the \textit{candidate classification} task, it is interesting too 
see how well non-expert humans can do it.

To this end we crowd-annotated\footnote{Using the Figure-Eight platform - \url{https://www.figure-eight.com/}} $250$ of the extracted candidate bigrams for whether or not they should be a WT, with each candidate annotated by $7$ annotators. The guidelines asked the annotators not to check their answer in Wikipedia, and explained that there are no wrong answers. On average, annotators achieved an accuracy of $0.76$ (std=$0.06$). Taking the majority vote for each candidate attains an accuracy of $0.81$. This suggests that individually, non-experts are better than the majority-class baseline, but not as good as the classification model, 
while the ``wisdom of the crowd" is on par with the latter.

Initially we were concerned that although we asked annotators not to look in Wikipedia for the answers, they nonetheless will do so. Conversely, with no answers being considered wrong, and no test questions, one might be concerned that annotators would answer at random to quickly collect their pay and move on to the next task. To alleviate these concerns we published the task in a special channel, 
whose annotators have proven trustworthy in past tasks. Furthermore, the mediocre accuracy, and a mediocre mean inter-annotator Cohen's Kappa of $0.47$, suggests that neither of these concerns turned out to be a major issue. On the one hand, these values are not high enough to suggest that annotators verified their work via Wikipedia, and on the other, they \textit{are} high enough to suggest that at least most annotators tried to answer in earnest.

\subsection{Application to Domain-Specific ATE}\label{sec:domain-spec}

Can domain-independent ATE be useful for domain-specific ATE? To explore the inter-relations between the two we considered the \textit{ACL RD-TEC 2.0} ATE benchmark \cite{qasemizadeh2016acl}, from which we extracted \textit{all}\footnote{No frequency 
threshold was set since many terms appear only once.} noun phrase bigrams and deduced their label 
as described in Section \ref{sec:data}. In total, $2139$ bigrams were identified in this dataset, of which $1134$ ($53\%$) were implied to be DSTs. We then searched for these bigrams in Wikipedia, finding $310$ of them therein, and asked whether being a WT is indicative for being a DST. 

This analysis suggest that being a WT is indeed
a strong indicator for being a DST: $80\%$ of the WTs are 
DSTs.\footnote{A cursory examination of the remaining $20\%$ suggests that one reason for not being a DST is that the WT is an AWT, and another is that the bigram is part of a longer n-gram, which \textit{is} labeled as a DST. We defer a more careful analysis to future work.} 
Yet, while the set of bigrams identified in Wikipedia has a high precision, 
it is relatively small, and thus inferring all bigrams outside this set to be non-DSTs yields low accuracy ($0.56$), due to low recall.

Can a classifier which was trained for domain-independent ATE be useful for domain-specific ATE? To test this, we used the fine-tuned BERT model described in Section \ref{sec:cand_class} to predict which bigrams are DSTs, and attained an accuracy of $0.64$. While this falls short of its success on identifying WTs, it does provide a clear advantage over the $0.53$ accuracy of the majority-class baseline. This suggests that there may be some common linguistic characteristic to WTs and DSTs, some of which were captured by the fine-tuned BERT model. Indeed, if the model is trained on all bigrams extracted from LNCS, rather than just the ones in the train set of Section \ref{sec:cand_class}, the accuracy further 
increases to $0.71$. Error analysis of these predictions appears in the appendix.

%% file: discussion.tex
\section{Discussion and Future Work}

This work considers a domain-independent variant of the classical Automatic Term Extraction task. This makes relevant much larger corpora than those commonly used for domain-specific ATE, and a much more comprehensive evaluation benchmark induced by Wikipedia. It also circumvents one of the main problems in ATE - that of determining \textit{termhood} \cite{kageura1996methods}, i.e. whether or not a term is relevant to the domain - and allows focusing on other aspects of the task. Accordingly, we address the tasks of \textit{ambiguity detection} and \textit{surface forms clustering}, which did not receive much attention in previous works.

As far as we know, although detecting and solving ambiguity has been the subject of much research, determining whether a WT is an AWT is in fact a novel task.
Moreover, the clustering 
task emerging from our premise 
is somewhat uncommon; 
cluster analysis is usually applied with the goal of understanding the structure of large data by clustering it down to a manageable number of clusters.
Conversely, here, clustering is a mean to an end, rather than an analysis tool. Even though the number of items to cluster is very large, the desired clustering is of numerous, small-sized clusters. As clustering algorithms and evaluation techniques have traditionally been developed in the former setting, it may be interesting to more carefully understand their applicability to this one, and perhaps learn if and how they should be adapted.

The massive corpora available for domain-independent ATE carries more 
potential, in terms of scale, than was realized here. We used the smaller LNCS as our starting point, and extracted a moderate number of candidates, to allow rapid explorations of the various baselines described above. However, given time and resources, 
one could apply the suggested pipeline to the entire LNC, as we hope to do in future work. Such an endeavor would allow not only addressing questions of accuracy, but also of recall, that is, identifying which parts of Wikipedia are indeed reflected in LNC, and which are not. That is, while we consider our setting as domain-independent, our error analysis suggests that perhaps a better description would be multi-domain, or a mixture of domains, as future work might hopefully reveal.

Extracting a larger number of candidates would also require more careful filtering rules.
When analyzing \textit{candidate extraction}, the only processing of candidate texts was removal of stop-words. Then, in the supervised learning experiments of the two subsequent steps, further filtering was done, by keeping only one candidate from among a pair of candidates being a singular and plural form of one another. Finally, in the clustering step, and the augmentation of the data with many co-redirects, the more elaborate RBC rules were introduced. Scaling up the pipeline would require similar rules used already at the extraction stage.

In classifying WTs vs non-WTs and AWTs vs NAWTs we have relied on BERT, since it is easy to use and readily available. Hopefully, since it was trained on such a large number of examples, only part of which come from Wikipedia, that it is not too biased toward the latter. We have tried to control for that by analyzing bigram and trigram frequency in Wikipedia, and by demonstrating that the fine-tuned BERT performs well also on the unrelated benchmark of \citet{qasemizadeh2016acl}. Nonetheless, future work should do away with this 
potential 
dependency on Wikipedia.

Finally, we gave only cursory consideration to the first and last steps of the pipeline. Especially in light of our goal to expand the scale on which the pipeline operates, these steps should receive more careful attention in the future.

%% file: main.bbl
\begin{thebibliography}{33}
\expandafter\ifx\csname natexlab\endcsname\relax\def\natexlab#1{#1}\fi

\bibitem[{Amig{\'o} et~al.(2009)Amig{\'o}, Gonzalo, Artiles, and
  Verdejo}]{Amigo2009}
Enrique Amig{\'o}, Julio Gonzalo, Javier Artiles, and Felisa Verdejo. 2009.
\newblock \href {https://doi.org/10.1007/s10791-008-9066-8} {A comparison of
  extrinsic clustering evaluation metrics based on formal constraints}.
\newblock \emph{Information Retrieval}, 12(4):461--486.

\bibitem[{Auger et~al.(1991)Auger, Drouin et~al.}]{auger1991automatisation}
Pierre Auger, Patrick Drouin, et~al. 1991.
\newblock Automatisation des proc{\'e}dures de travail en terminographie.
\newblock \emph{Meta: journal des traducteurs/Meta: Translators' Journal},
  36(1):121--127.

\bibitem[{Bawakid(2015)}]{bawakid2015using}
Abdullah Bawakid. 2015.
\newblock Using wikipedia categories for discovering the themes of text
  documents.
\newblock In \emph{2015 7th International Conference on Intelligent
  Human-Machine Systems and Cybernetics}, volume~1, pages 452--455. IEEE.

\bibitem[{Bourigault(1992)}]{bourigault1992surface}
Didier Bourigault. 1992.
\newblock Surface grammatical analysis for the extraction of terminological
  noun phrases.
\newblock In \emph{Proceedings of the 14th conference on Computational
  linguistics-Volume 3}, pages 977--981. Association for Computational
  Linguistics.

\bibitem[{Brewster et~al.(2007)Brewster, Iria, Zhang, Ciravegna, Guthrie, and
  Wilks}]{brewster2007dynamic}
Christopher Brewster, Jos{\'e} Iria, Ziqi Zhang, Fabio Ciravegna, Louise
  Guthrie, and Yorick Wilks. 2007.
\newblock Dynamic iterative ontology learning.
\newblock \emph{Recent Advances in Natural Language Processing (RANLP 07)}.

\bibitem[{Castellv{\'\i} et~al.(2001)Castellv{\'\i}, Bagot, and
  Palatresi}]{castellvi2001automatic}
M~Teresa~Cabr{\'e} Castellv{\'\i}, Rosa~Estopa Bagot, and Jordi~Vivaldi
  Palatresi. 2001.
\newblock Automatic term detection: A review of current systems.
\newblock \emph{Recent advances in computational terminology}, 2:53--88.

\bibitem[{Chisholm et~al.(2016)Chisholm, Radford, and
  Hachey}]{chisholm2016discovering}
Andrew Chisholm, Will Radford, and Ben Hachey. 2016.
\newblock Discovering entity knowledge bases on the web.
\newblock In \emph{Proceedings of the 5th Workshop on Automated Knowledge Base
  Construction}, pages 7--11.

\bibitem[{Conrado et~al.(2013)Conrado, Pardo, and Rezende}]{conrado2013machine}
Merley Conrado, Thiago Pardo, and Solange Rezende. 2013.
\newblock A machine learning approach to automatic term extraction using a rich
  feature set.
\newblock In \emph{Proceedings of the 2013 NAACL HLT Student Research
  Workshop}, pages 16--23.

\bibitem[{Fraser et~al.(2019)Fraser, Nejadgholi, De~Bruijn, Li, LaPlante, and
  Abidine}]{fraser2019extracting}
Kathleen~C Fraser, Isar Nejadgholi, Berry De~Bruijn, Muqun Li, Astha LaPlante,
  and Khaldoun Zine~El Abidine. 2019.
\newblock Extracting umls concepts from medical text using general and
  domain-specific deep learning models.
\newblock \emph{arXiv preprint arXiv:1910.01274}.

\bibitem[{Gupta and Manning(2011)}]{gupta2011analyzing}
Sonal Gupta and Christopher~D Manning. 2011.
\newblock Analyzing the dynamics of research by extracting key aspects of
  scientific papers.
\newblock In \emph{Proceedings of 5th international joint conference on natural
  language processing}, pages 1--9.

\bibitem[{Hartigan(1975)}]{hartigan1975}
John~A. Hartigan. 1975.
\newblock \emph{Clustering Algorithms}, 99th edition.
\newblock John Wiley \& Sons, Inc., New York, NY, USA.

\bibitem[{Honnibal and Johnson(2015)}]{honnibal2015improved}
Matthew Honnibal and Mark Johnson. 2015.
\newblock An improved non-monotonic transition system for dependency parsing.
\newblock In \emph{Proceedings of the 2015 Conference on Empirical Methods in
  Natural Language Processing}, pages 1373--1378.

\bibitem[{Kageura and Umino(1996)}]{kageura1996methods}
Kyo Kageura and Bin Umino. 1996.
\newblock Methods of automatic term recognition: A review.
\newblock \emph{Terminology. International Journal of Theoretical and Applied
  Issues in Specialized Communication}, 3(2):259--289.

\bibitem[{Kim et~al.(2003)Kim, Ohta, Tateisi, and Tsujii}]{kim2003genia}
J-D Kim, Tomoko Ohta, Yuka Tateisi, and Jun’ichi Tsujii. 2003.
\newblock Genia corpus—a semantically annotated corpus for bio-textmining.
\newblock \emph{Bioinformatics}, 19(suppl\_1):i180--i182.

\bibitem[{Lewoniewski et~al.(2016)Lewoniewski, Wecel, and
  Abramowicz}]{lewoniewski2016quality}
W{\l}odzimierz Lewoniewski, Krzysztof Wecel, and Witold Abramowicz. 2016.
\newblock Quality and importance of wikipedia articles in different languages.
\newblock In \emph{International Conference on Information and Software
  Technologies}, pages 613--624. Springer.

\bibitem[{Lossio-Ventura et~al.(2016)Lossio-Ventura, Jonquet, Roche, and
  Teisseire}]{lossio2016biomedical}
Juan~Antonio Lossio-Ventura, Clement Jonquet, Mathieu Roche, and Maguelonne
  Teisseire. 2016.
\newblock Biomedical term extraction: overview and a new methodology.
\newblock \emph{Information Retrieval Journal}, 19(1-2):59--99.

\bibitem[{Luhn(1957)}]{luhn1957statistical}
Hans~Peter Luhn. 1957.
\newblock A statistical approach to mechanized encoding and searching of
  literary information.
\newblock \emph{IBM Journal of research and development}, 1(4):309--317.

\bibitem[{Mohapatra et~al.(2018)Mohapatra, Deng, Gupta, Dasgupta, Paradkar,
  Mahindru, Rosu, Tao, and Aggarwal}]{mohapatra2018domain}
Prateeti Mohapatra, Yu~Deng, Abhirut Gupta, Gargi Dasgupta, Amit Paradkar,
  Ruchi Mahindru, Daniela Rosu, Shu Tao, and Pooja Aggarwal. 2018.
\newblock Domain knowledge driven key term extraction for it services.
\newblock In \emph{International Conference on Service-Oriented Computing},
  pages 489--504. Springer.

\bibitem[{Nakagawa and Mori(2002)}]{nakagawa2002simple}
Hiroshi Nakagawa and Tatsunori Mori. 2002.
\newblock A simple but powerful automatic term extraction method.
\newblock In \emph{COLING-02 on COMPUTERM 2002: second international workshop
  on computational terminology-Volume 14}, pages 1--7. Association for
  Computational Linguistics.

\bibitem[{Peng et~al.(2016)Peng, Song, and Roth}]{peng2016event}
Haoruo Peng, Yangqiu Song, and Dan Roth. 2016.
\newblock Event detection and co-reference with minimal supervision.
\newblock In \emph{Proceedings of the 2016 conference on empirical methods in
  natural language processing}, pages 392--402.

\bibitem[{Pennington et~al.(2014)Pennington, Socher, and
  Manning}]{pennington2014glove}
Jeffrey Pennington, Richard Socher, and Christopher Manning. 2014.
\newblock Glove: Global vectors for word representation.
\newblock In \emph{Proceedings of the 2014 conference on empirical methods in
  natural language processing (EMNLP)}, pages 1532--1543.

\bibitem[{Plante and Dumas(1998)}]{plante1998depoulliment}
Pierre Plante and Lucie Dumas. 1998.
\newblock Le d{\'e}poulliment terminologique assist{\'e} par ordinateur.
\newblock \emph{Terminogramme}, 46:24--28.

\bibitem[{QasemiZadeh and Schumann(2016)}]{qasemizadeh2016acl}
Behrang QasemiZadeh and Anne-Kathrin Schumann. 2016.
\newblock The acl rd-tec 2.0: A language resource for evaluating term
  extraction and entity recognition methods.
\newblock In \emph{Proceedings of the Tenth International Conference on
  Language Resources and Evaluation (LREC'16)}, pages 1862--1868.

\bibitem[{Slonim et~al.(2005)Slonim, Atwal, Tka{\v c}ik, and
  Bialek}]{Slonim18297}
Noam Slonim, Gurinder~Singh Atwal, Ga{\v s}per Tka{\v c}ik, and William Bialek.
  2005.
\newblock \href {https://doi.org/10.1073/pnas.0507432102} {Information-based
  clustering}.
\newblock \emph{PNAS}, 102(51):18297--18302.

\bibitem[{Slonim et~al.(2002)Slonim, Friedman, and Tishby}]{Slonim2002}
Noam Slonim, Nir Friedman, and Naftali Tishby. 2002.
\newblock \href {https://doi.org/10.1145/564376.564401} {Unsupervised document
  classification using sequential information maximization}.
\newblock In \emph{Proceedings of the 25th Annual International ACM SIGIR
  Conference on Research and Development in Information Retrieval}, SIGIR '02,
  pages 129--136, New York, NY, USA. ACM.

\bibitem[{Thalhammer and Rettinger(2016)}]{thalhammer2016pagerank}
Andreas Thalhammer and Achim Rettinger. 2016.
\newblock Pagerank on wikipedia: towards general importance scores for
  entities.
\newblock In \emph{European Semantic Web Conference}, pages 227--240. Springer.

\bibitem[{Tsai et~al.(2013)Tsai, Kundu, and Roth}]{tsai2013concept}
Chen-Tse Tsai, Gourab Kundu, and Dan Roth. 2013.
\newblock Concept-based analysis of scientific literature.
\newblock In \emph{Proceedings of the 22nd ACM international conference on
  Information \& Knowledge Management}, pages 1733--1738. ACM.

\bibitem[{Usbeck et~al.(2015)Usbeck, R{\"o}der, Ngonga~Ngomo, Baron, Both,
  Br{\"u}mmer, Ceccarelli, Cornolti, Cherix, Eickmann
  et~al.}]{usbeck2015gerbil}
Ricardo Usbeck, Michael R{\"o}der, Axel-Cyrille Ngonga~Ngomo, Ciro Baron,
  Andreas Both, Martin Br{\"u}mmer, Diego Ceccarelli, Marco Cornolti, Didier
  Cherix, Bernd Eickmann, et~al. 2015.
\newblock Gerbil: general entity annotator benchmarking framework.
\newblock In \emph{Proceedings of the 24th international conference on World
  Wide Web}, pages 1133--1143. International World Wide Web Conferences
  Steering Committee.

\bibitem[{Velardi et~al.(2008)Velardi, Navigli, and
  Pierluigi}]{velardi2008mining}
Paola Velardi, Roberto Navigli, and D~Pierluigi. 2008.
\newblock Mining the web to create specialized glossaries.
\newblock \emph{IEEE Intelligent Systems}, 23(5):18--25.

\bibitem[{Zaheer et~al.(2017)Zaheer, Kottur, Ravanbakhsh, Poczos,
  Salakhutdinov, and Smola}]{zaheer2017deep}
Manzil Zaheer, Satwik Kottur, Siamak Ravanbakhsh, Barnabas Poczos, Russ~R
  Salakhutdinov, and Alexander~J Smola. 2017.
\newblock Deep sets.
\newblock In \emph{Advances in neural information processing systems}, pages
  3391--3401.

\bibitem[{Zhang et~al.(2016)Zhang, Gao, and Ciravegna}]{zhang2016jate}
Ziqi Zhang, Jie Gao, and Fabio Ciravegna. 2016.
\newblock Jate 2.0: Java automatic term extraction with apache solr.
\newblock In \emph{Proceedings of the Tenth International Conference on
  Language Resources and Evaluation (LREC'16)}, pages 2262--2269.

\bibitem[{Zhang et~al.(2008)Zhang, Iria, Brewster, and
  Ciravegna}]{zhang2008comparative}
Ziqi Zhang, Jos{\'e} Iria, Christopher Brewster, and Fabio Ciravegna. 2008.
\newblock A comparative evaluation of term recognition algorithms.
\newblock In \emph{LREC}, volume~5.

\bibitem[{Zhang et~al.(2018)Zhang, Petrak, and Maynard}]{zhang2018adapted}
Ziqi Zhang, Johann Petrak, and Diana Maynard. 2018.
\newblock Adapted textrank for term extraction: a generic method of improving
  automatic term extraction algorithms.
\newblock \emph{Procedia Computer Science}, 137:102--108.

\end{thebibliography}
